\documentclass[11pt,a4paper]{article}
\usepackage{authblk}
\usepackage[hyperref]{acl2019}
\usepackage{times}
\usepackage{latexsym}
\usepackage{multirow, makecell}
\usepackage{textcomp}
\usepackage{url}
\usepackage{environ}
\usepackage{tabularx}
\usepackage{enumitem}
\usepackage{graphicx}
\usepackage{todonotes}
\usepackage{color, colortbl}
\usepackage{listings}
\newsavebox{\mybox}

\definecolor{Gray}{gray}{0.9}

\newcommand{\red}[1]{\textcolor{red}{#1}}
\newcommand{\blue}[1]{\textcolor{blue}{#1}}
\newcommand{\orange}[1]{\textcolor{orange}{#1}}

\newcommand{\printfnsymbol}{%
  \textsuperscript{*}%
}

\aclfinalcopy % Uncomment this line for the final submission
\def\aclpaperid{2616} %  Enter the acl Paper ID here

\title{Constrained Decoding for Neural NLG from Compositional Representations in Task-Oriented Dialogue}

\author{\textbf{Anusha Balakrishnan\thanks{~~Alphabetical by first name} }}
\author{\textbf{Jinfeng Rao}\printfnsymbol}
\author{\textbf{Kartikeya Upasani}\printfnsymbol}
\author{\textbf{Michael White}\printfnsymbol\thanks{~~Work done while on leave from Ohio State University} }
\author{\textbf{Rajen Subba}\printfnsymbol}

\affil[]{Facebook Conversational AI}
\affil[]{\texttt{\{anushabala,raojinfeng,kart,mwhite14850,rasubba\}@fb.com}}

\date{}

\begin{document}
\maketitle
\begin{abstract}
Generating fluent natural language responses from structured semantic representations is a critical step in task-oriented conversational systems. Avenues like the E2E NLG Challenge have encouraged the development of neural approaches, particularly sequence-to-sequence (Seq2Seq) models for this problem. The semantic representations used, however, are often underspecified, which places a higher burden on the generation model for sentence planning, and also limits the extent to which generated responses can be controlled in a live system. In this paper, we (1) propose using 
%tree-structured semantic representations, like those used in traditional rule-based %NLG systems, with Seq2Seq models; (2) introduce a challenging dataset using this 
tree-structured semantic representations, like those used in traditional rule-based NLG systems, for better discourse-level structuring and sentence-level planning; (2) introduce a challenging dataset using this
representation for the weather domain; (3) introduce a constrained decoding approach for Seq2Seq models that leverages this representation to improve semantic correctness; and (4) demonstrate promising results on our dataset and the E2E dataset.
\end{abstract}

\section{Introduction}\label{sec:intro}

Generating fluent natural language responses from structured semantic representations is a critical step in task-oriented conversational systems. With their end-to-end trainability, neural approaches to natural language generation (NNLG), particularly sequence-to-sequence (Seq2Seq) models, have been promoted with great fanfare in recent years \citep{Wen-etal-D15-1199,Wen-etal-N16-1015,Mei-Bansal-Walter-N16-1086, D16-1032,DandJ-P16-2008}, and avenues like the recent E2E NLG challenge \citep{W18-6539,e2echallengeoverview} have made available large datasets to promote the development of these models.  Nevertheless, current NNLG models arguably remain inadequate for most real-world task-oriented dialogue systems, given their inability to (i) reliably perform common sentence planning and discourse structuring operations \cite{Reed-et-al:W18-6535}, (ii) generalize to complex inputs \cite{D17-1239}, and (3) avoid generating texts with semantic errors including hallucinated content \citep{W18-6539,e2echallengeoverview}.\footnote{Also see \url{https://ehudreiter.com/2018/11/12/hallucination-in-neural-nlg/}.}

In this paper, we explore the extent to which these issues can be addressed by incorporating lessons from pre-neural NLG systems into a neural framework. We begin by arguing in favor of enriching the input to neural generators to include discourse relations --- long taken to be central in traditional NLG --- and underscore the importance of exerting control over these relations when generating text, particularly when using user models to structure responses. In a closely related work, \citet{Reed-et-al:W18-6535}, the authors add control tokens (to indicate contrast and sentence structure) to a flat input MR, and show that these can be effectively used to control structure. However, their methods are only able to control the presence or absence of these relations, without more fine-grained control over their structure. We thus go beyond their approach and propose using full tree structures as inputs, and generating tree-structured outputs as well. This allows us to define a novel method of constrained decoding for standard sequence-to-sequence models for generation, which helps ensure that the generated text contains all and only the specified content, as in classic approaches to surface realization. 

On the E2E dataset, our experiments demonstrate much better control over \textsc{Contrast} relations than using Reed et al.'s method, and also show improved diversity and expressiveness over standard baselines. We also release a new dataset of responses in the weather domain, which includes the \textsc{Justify}, \textsc{Join} and \textsc{Contrast} relations, and where discourse-level structures come into play. On both E2E and weather datasets, we show that constrained decoding over our enriched inputs results in higher semantic correctness as well as better generalizability and data efficiency. 

The rest of this paper is organized as follows: Section~\ref{sec:compositional} describes the motivation for using compositional inputs organized around discourse relations. Section~\ref{sec:data} explains our data collection approach and dataset.\footnote{The datasets and implementations can be found at \url{https://github.com/facebookresearch/TreeNLG}.} Section~\ref{sec:model} shows how to incorporate compositional inputs into NNLG and describes our constrained decoding algorithm. Section~\ref{sec:expt} presents our experimental setup and results.

% \begin{itemize}[leftmargin=*,nolistsep,noitemsep]
% \item We demonstrate for the first time how compositional inputs organized around discourse relations can be used in NNLG for task-oriented dialogue (Section~\ref{sec:compositional});

% \item We release a new dataset for the weather domain and a version of the E2E dataset that make use of compositional inputs (Section~\ref{sec:data}); 

% \item We define a novel method of constrained decoding that helps ensure that all and only the specified content is realized and release an open source implementation (Section~\ref{sec:model}); and

% \item We demonstrate experimentally, using both automatic measures and human evaluation on both datasets, that our methods enable much greater control over desired outputs than previous methods, while also enhancing data efficiency and generalizability (Section~\ref{sec:expt}).
% \end{itemize}
% In summary, the contributions of the paper are: (1) We demonstrate for the first time how compositional inputs organized around discourse relations can be used in NNLG for task-oriented dialogue (Section~\ref{sec:compositional}). (2) We define a novel method of constrained decoding that helps ensure semantic correctness (Section~\ref{sec:model}). (3) We demonstrate experimentally, using automatic measures and human evaluation, that our methods enable much greater control over desired outputs than previous methods, while also enhancing data efficiency and generalizability (Section~\ref{sec:expt}).

\begin{table}[t!]
\begin{center}
\small
\begin{tabular}{l p{5cm}}
\hline 
\textbf{Reference 1} & JJ's Pub is not family friendly, but has a high customer rating of 5 out of 5. It is a restaurant near the Crowne Plaza Hotel. \\
\hline 
\textbf{Reference 2} & JJ's Pub is not a family friendly restaurant. It has a high customer rating of 5 out of 5. You can find it near the Crowne Plaza Hotel. \\
\hline
\textbf{E2E MR} & name[JJ's Pub] rating[5 out of 5] familyFriendly[no] eatType[restaurant] near[Crowne Plaza Hotel] \\ 
\hline
\multirowcell{8}{\textbf{Our MR for} \\ \textbf{Reference 1}} & \textbf{CONTRAST} [ \\
 &   \hspace{5mm}\textbf{INFORM} [ name[JJ's Pub] \\
 &   \hspace{5mm} \hspace{14mm} familyFriendly[no] ] \\
 &   \hspace{5mm}\textbf{INFORM} [ rating[5 out of 5] ] ]\\
 %&   ] \\
 &   \textbf{INFORM} [ \\
 &   \hspace{5mm}eatType[restaurant] \\
 &   \hspace{5mm}near[Crowne Plaza Hotel] ]\\
 %&   ] \\
\hline
\end{tabular}
\end{center}
\caption{\label{tab:mr-example} Sample reference responses, their corresponding meaning representation in the E2E dataset, and its MR according to our proposed ontology. }
\end{table}

\section{Towards More Expressive Meaning Representations} \label{sec:compositional}

\subsection{Limitations of Flat MRs}

In the E2E dataset, meaning representations (MRs) are a flat list of key-value pairs, where each key is a slot name that needs to be mentioned, and the value is the value of that slot (see Table \ref{tab:mr-example}). In \citet{Wen-etal-D15-1199}, MRs have a similar structure, and additionally contain information about the \textbf{dialog act} that needs to be conveyed (\textsc{Request}, \textsc{Inform}, etc.). These MRs are sufficient to capture basic semantic information, but fail to capture rhetorical (or discourse) relations, like \textsc{Contrast}, that have long been taken to be central to generating coherent discourse in traditional NLG \citep{mann-thompson:88,discoursemotivationpersuasion,Reiter-Dale:2000,discourseactmotivations2usermodeling}.   The two references in Table \ref{tab:mr-example} illustrate this problem with the expressiveness of such flat MRs. Critical discourse information, like whether two attributes should be contrasted (or whether to justify a recommendation, etc.), is not captured by the MR. This poses a dual challenge: First, since the MR does not specify these discourse relations, crowdworkers creating the dataset in turn have no instructions on when to use them, and must thus use their own judgment in creating a natural-sounding response. While the E2E organizers tout the resulting response variations as a plus, \citet{Reed-et-al:W18-6535} find that current neural systems are unable to learn to express discourse relations effectively with this dataset, and explore ways of enriching input MRs to do so. Indeed, now that the E2E system outputs have been released, a search through outputs from all participating systems reveals only 43 outputs (0.4\% out of 10080) containing contrastive tokens, on a test set containing about 300 contrastive samples.\footnote{An additional 86 outputs contained these tokens, but were generated by the \texttt{TR2} template-based system \citep{thomsonreuterstemplatee2e}. The expected number of contrastive system outputs would be 4,200 if each of the 14 participating systems produced contrastive tokens consistently with the data distribution.}

Second, going beyond Reed et al., we argue that the \textbf{controllability} of these relations through MRs is desirable in live conversational systems, where external knowledge like user models may inform decisions around contrast, grouping, or justifications. While several studies have shown that controlling such discourse behaviors can be critical to user perceptions of quality and naturalness \cite{discourseactmotivations1usermodeling,Carenini-Moore:2006,walker-etal:07,White-Clark-Moore:2010,demberg2011strategy}, flat MRs provide no means to do so.  This leaves it to the neural model to learn general trends in the data, such as contrasting a good attribute like a 5-star rating with a typically dispreferred attribute like not being family friendly or serving English food.  However, sometimes people are interested in adult-oriented establishments, and some people may even like English food; for users with these preferences, text generated according to general trends will be incoherent. For example, for a user known to be seeking an adult-oriented locale, Ref.~1 in Table~\ref{tab:mr-example} would be incoherent, and less preferable than a non-contrastive alternative such as \textit{JJ's Pub is a highly-rated restaurant for adults near the Crowne Plaza Hotel}.

\begin{table*}[t]
\begin{center}
\small
\resizebox{0.8\textwidth}{!}{%
\begin{tabularx}{\linewidth}{l p{14.5cm}}
\hline 
\textbf{Reference} &  It'll be sunny throughout this weekend. The high will be in the 60s, but expect temperatures to drop as low as 43 degrees by Sunday evening. There's also a chance of strong winds on Saturday morning. \\
\hline
\textbf{Flat MR} &  \texttt{condition1[sunny] date\_time1[this weekend] avg\_high1[60s] low2[43] date\_time2[Sunday evening] chance3[likely] wind\_summary3[strong] date\_time3[Saturday morning]}\\ 
\hline
\multirow{6}{*}{\textbf{Our MR}} & \texttt{\blue{\textbf{INFORM}} [ condition[sunny], \orange{date\_time\_range[} colloquial[this weekend ] \orange{]} \blue{]} }\\
 &   \texttt{\blue{\textbf{CONTRAST}} [} \\
 & \texttt{\hspace{3mm} \orange{\textbf{INFORM}} [  avg\_high[60s] \red{date\_time[} [colloquial this weekend ] \red{]} \orange{]}} \\
 & \texttt{\hspace{3mm} \orange{\textbf{INFORM}} [ low[43] \red{date\_time[} week\_day[Sunday] colloquial[evening] \red{]} \orange{]}} \\
 & \texttt{\blue{]}}  \\
 & \texttt{\blue{\textbf{INFORM}} [ chance[likely], wind\_summary[heavy], \orange{date\_time[} week\_day[Saturday] colloquial[morning] \orange{]} \blue{]}}\\

\hline
\end{tabularx}}
\end{center}
\caption{\label{fig:mrcomparisons} Sample flat MR with reference compared against our proposed tree-structured MR. Nodes in blue are all children of the root node of the tree.}
\vspace{-0.4cm}
\end{table*}

\subsection{Tree-Structured MRs}

In order to overcome these challenges, we propose the use of structured meaning representations like those explored widely in (hybrid) rule-based NLG systems \citep{nlgoverviewrulebasedRambowWalker,Reiter-Dale:2000,walker-etal:07}. Our representation consists of three parts:

\begin{enumerate}[noitemsep,nolistsep]
    \item \textbf{Argument} can be any entity or slot mentioned in a response, like the name of a restaurant or the date. Some arguments can be complex and contain sub-arguments (e.g. a \texttt{date\_time} argument has subfields like \texttt{week\_day} and \texttt{month}). 
    %Examples: \texttt{date\_time}, \texttt{customerrating}, \texttt{temp\_high}.
    \item \textbf{Dialog act} is an atomic unit that could correspond linguistically to a single clause. A dialog act can contain one or more arguments that need to be expressed. Examples: \textsc{Inform}, \textsc{Yes}, \textsc{Recommend}.
    \item \textbf{Discourse relation} defines the relationships between dialog acts. A single discourse relation may contain multiple other dialog or discourse relations, allowing for potentially arbitrary degrees of nesting. Examples: \textsc{Join}, \textsc{Justify}, \textsc{Contrast}.
\end{enumerate}

A meaning representation that uses this formulation can consist of an arbitrary number and combination of discourse relations and dialog acts, resulting in a nested tree-structured MR with much higher expressiveness and specificity. Table \ref{tab:mr-example}, seen earlier, shows an example of an MR structured in this way, as well as the corresponding ``flat'' MR and its reference in the E2E dataset.

%Since discourse relations can contain arbitrary nestings of other discourse relations and dialog acts, this results in a tree-structured MR. Table \ref{tab:mr-example} shows an example of an MR structured in this way, as well as the corresponding ``flat'' MR and its reference in the E2E dataset. \todo{This figure reference seems out of date, it comes much later and doesn't show a flat MR??}

In addition to improved expressiveness, this representation results in more atomic definitions of dialog acts and arguments than in flat MRs. For example, consider the example in the weather domain from Table~\ref{fig:mrcomparisons}:  The response contains multiple dialog acts, a contrast and several instances of ellipsis and grouping (i.e., temperatures are grouped and mentioned separately from wind condition). Additionally, some arguments, like \texttt{date\_time}, occur multiple times in the response and correspond to different dialog acts, with several different values. A flat MR will struggle to represent 1) the correspondence of arguments to dialog acts; 2) what attributes to group and contrast and 3) semantic equivalence of arguments like \texttt{date\_time1} and \texttt{date\_time2}. On the other hand, our MRs ease discourse-level learning and encourage reuse of arguments across multiple dialog acts.

\section{Dataset}\label{sec:data} 
%\todo{need to talk more about why weather domain? } 
With this representation in mind, we created an ontology of dialog acts, discourse relations, and arguments, for the weather domain. 
%\todo{(talk about \texttt{task} args and errors)} 
Our motivation for choosing the weather domain, as explored in \cite{liang2009learning}, is that this domain offers significant complexity for NLG. Weather forecast summaries in particular can be very long, and require reasoning over several disjoint pieces of information. In this work, we focused on collecting a dataset that showcases the complexity of weather summaries over date/time ranges. Our weather dataset is also unique in that it was collected in a \textit{conversational} setup (see below).

%\begin{table*}[t]
%\centering
%\tiny
%\begin{tabularx}{\linewidth}{l| l | l l l l}
%\textbf{Dialog Acts} & \textbf{Discourse relations} & \textbf{Arguments} &  &  & \vspace{2mm}\\
%\hline 
%INFORM & \textsc{JOIN} & attire\textsuperscript{[n]} & activity\textsuperscript{[n]} & condition\textsuperscript{[n]} & humidity\textsuperscript{[n]}\\
%RECOMMEND & \textsc{JUSTIFY} & precip\_amount & precip\_amount\_unit & precip\_chance & precip\_chance\_summary \\
%YES & \textsc{CONTRAST} & precip\_type & sunrise\_time & sunset\_time & temp \\
%NO &  & temp\_high\textsuperscript{[s]} & temp\_low\textsuperscript{[s]} & temp\_unit & wind\_speed\textsuperscript{[n]}  \\
%ERROR & & wind\_speed\_unit & bad\_arg & bad\_value & error\_reason \\
% & & task & date\_time* & date\_time\_range* & location*

%\end{tabularx}
%\caption{Ontology for the weather domain dataset that we collected. Arguments marked with * are nested arguments (see Table \ref{tab:subfields}). \textsuperscript{[n]} indicates arguments that have a corresponding \texttt{\_not} argument; \textsuperscript{[s]} indicates arguments that have a corresponding \texttt{\_summary}.}
%\label{tab:weatherontology}

%\begin{tabularx}{\linewidth}{l l}
%\textbf{Argument} & \textbf{Subfields}\\
%\hline 
%date\_time & year, month, day, weekday, colloquial \\
%date\_time\_range & start\_year, start\_month, start\_day, start\_weekday, end\_year, end\_month, end\_day, end\_weekday, colloquial \\
%location & city, region, country, colloquial \\
%\end{tabularx}
%\caption{Defined subfields for nested arguments in our ontology.}
%\label{tab:subfields}
%\end{table*}

We collected our dataset in multiple stages:
%\begin{enumerate}[noitemsep,nolistsep,leftmargin=*]

\textbf{1. Query collection.} We asked crowdworkers to come up with sample queries in the weather domain, like \textit{What's the weather like tomorrow?} and \textit{Do I need an umbrella tonight?} 

\textbf{2. Query annotation.} We then wrote rules to automatically parse these queries, and extract key pieces of information, like the location, date, and any attributes that the user specifically requested in the question. 
    
\textbf{3. MR generation}. Our goal was to create MRs that are sufficiently expressive and straightforward to create automatically in a practical system. In the weather domain, it's conceivable that the NLG system has access to a weather API that provides it with detailed weather forecasts for the range requested by the user. To mimic this setting, we generated artificial weather forecasts for every user query based on the arguments (full argument set in Table~\ref{tab:weatherontology}) in the user query. We then created the tree-structured MR by applying a few different types of automatic rules, like adding \texttt{CONTRAST} to weather conditions that are in opposition. We add more details of our response generation method and the specific rules for MR creation in Appendix \ref{sec:scengen} and \ref{sec:mrcreation}. 
 
\textbf{4. Response generation and annotation.} We presented these tree-structured MRs to trained annotators, and asked them to write responses that expressed the MRs. They were also given the user query and asked to make their responses natural given the query. They were allowed to elide information when arguments were repeated across dialog acts, and could choose the most appropriate surface forms for any arguments based on contextual clues (e.g. referring to a date as \textit{tomorrow}, rather than \textit{April 24\textsuperscript{th}}, depending on the user's date). Finally, we asked them to label response spans corresponding to each argument, dialog act, and discourse relation in the MR.

%\indent \textbf{4. Response generation and annotation.} We presented these prepared tree-structured MRs to trained annotators, and asked them to write responses that expressed the content of the MR. Annotators were presented with the user query to aid in writing responses that are more relevant to the query. In order to get a dataset with complete alignments, we also asked annotators to label spans in their response corresponding to each argument, dialog act, and discourse relation.
    
\textbf{5. Quality evaluation.} Finally, we presented a different group of annotators with the annotated responses, and asked them to provide evaluations of \textit{fluency}, \textit{correctness}, \textit{naturalness}, and \textit{annotation correctness}. 
% \todo{more info here}

\begin{table}[t]
\centering
\scriptsize
\begin{tabularx}{\linewidth}{l| l}
\hline
\textbf{Dialog Acts} & \texttt{INFORM}, \texttt{RECOMMEND}, \texttt{YES}, \texttt{NO}, \texttt{ERROR} \\
\hline
\textbf{Discourse Relations} & \textsc{Join}, \textsc{Contrast}, \textsc{Justify} \\
\hline
\multirow{7}{*}{\textbf{Arguments}} & date\_time*, date\_time\_range*, location* \\
& attire\textsuperscript{[n]}, activity\textsuperscript{[n]}, condition\textsuperscript{[n]}, humidity\textsuperscript{[n]} \\
& precip\_amount, precip\_amount\_unit, precip\_chance \\
& precip\_chance\_summary, precip\_type, sunrise\_time, \\
& temp, temp\_high\textsuperscript{[s]}, temp\_low\textsuperscript{[s]}, temp\_unit \\
& wind\_speed\textsuperscript{[n]}, wind\_speed\_unit, sunset\_time, task \\
& bad\_arg, bad\_value, error\_reason \\
\hline
\end{tabularx}
\caption{Ontology for the weather domain dataset that we collected. Arguments marked with * are nested arguments (see Table \ref{tab:subfields}). \textsuperscript{[n]} indicates arguments that have a corresponding \texttt{\_not} argument; \textsuperscript{[s]} indicates arguments that have a corresponding \texttt{\_summary}. \\}
\label{tab:weatherontology}
\vspace{-0.7cm}
\end{table}

\subsection{Dataset statistics}
%\todo{comparison to E2E here} 
Our final dataset has 33,493 examples. Each example comprises a user query, the synthetic user context (datetime and location), the tree-structured MR, the response, and a complete tree-structured annotation of the response. Table \ref{tab:weatherexamplerow} contains an example from our dataset; as shown, the response annotation structure closely mirrors that of the MR itself. The MRs and responses in the dataset range from very simple (a single dialog act) to very complex (an MR with a depth and width of 4). A distribution of this complexity is shown in Table~\ref{tab:freq-dist}. The vocabulary size is 1485, and the max/average/min lengths of responses are 151/40.6/8. The dataset also poses several challenges in addition to syntactic and semantic complexity. As mentioned before, it has a rich set of referring expressions for dates and date ranges. It also contains user queries on which the written response was based, thus creating the opportunity for studies on improving naturalness or relevance with respect to the user query. These could be useful in particular for learning to express recommendations and justifications, as well as \texttt{YES} and \texttt{NO} dialog acts.

Our final training set contains 25,390 examples, with 11,879 unique MRs. (We consider two MRs to be identical if they have the same \textbf{delexicalized} tree structure --- see Section \ref{subsec:delex}.) The test set contains 3,121 examples, of which 1.1K (35\%) have unique MRs \textit{that have never been seen in the training set}.  

\begin{table}[t]
\scriptsize
\begin{tabularx}{\linewidth}{l | l}
\hline
\textbf{Argument} & \textbf{Subfields}\\
\hline 
\textbf{date\_time} & year, month, day, weekday, colloquial \\
\multirow{2}{*}{\textbf{date\_time\_range}} & start\_year, start\_month, start\_day, start\_weekday \\
& end\_year, end\_month, end\_day, end\_weekday, colloquial \\
\textbf{location} & city, region, country, colloquial \\
\hline
\end{tabularx}
\caption{Defined subfields for nested arguments.}
\label{tab:subfields}
\vspace{-0.4cm}
\end{table}

\begin{table}
\centering
\scriptsize
\begin{tabularx}{\linewidth}{l| l l l l l l}
\textbf{Frequency} & \textbf{0} & \textbf{1} & \textbf{2} & \textbf{3} & \textbf{4} & \textbf{5} \\
\hline
\textbf{\# Dialog Acts} & 0 & 6469 & 12077 & 9801 & 4095 & 685 \\
\textbf{\# Discourse Rels} & 18137 & 12494 & 2393 & 103 & 1 & 0 \\
\hline
\end{tabularx}
\caption{Frequency distribution of number of dialog acts and discourse relations in the weather dataset.}
\label{tab:freq-dist}
\vspace{-0.6cm}
\end{table}

%\todo{add breakdown of act frequencies, etc.}
%\todo{diversity stats}

\begin{table*}%[]
    \small
    \centering
    \resizebox{0.7\textwidth}{!}{%
    \begin{tabularx}{\linewidth}{p{1.5cm}|p{1.5cm}|p{10cm}|p{4cm}}
        \textbf{Query} & \textbf{Context} & \textbf{MR} & \textbf{Response} \\
        \multirow{3}{=}{When will it snow next?} & \multirow{3}{=}{Reference date: 29th September 2018} & \red{[CONTRAST}  & \multirow{11}{=}{Parker is not expecting any snow, but today there's a very likely chance of heavy rain showers, and it'll be partly cloudy} \\ 
         & &\hspace{3mm}\orange{[INFORM\_1}  & \\
         & & \hspace{6mm} \blue{[LOCATION} [CITY Parker] \blue{]} [CONDITION\_NOT snow ] &  \\
         & & \hspace{6mm} \blue{[DATE\_TIME} [DAY 29] [MONTH September] [YEAR 2018] \blue{]} & \\
         & & \hspace{3mm} \orange{]} & \\
         & & \hspace{3mm}\orange{[INFORM\_2} &  \\ 
         & & \hspace{6mm} \blue{[DATE\_TIME} [DAY 29] [MONTH September] [YEAR 2018] \blue{]}  & \\
         & & \hspace{6mm} \blue{[LOCATION} [CITY Parker] \blue{]} & \\
         & & \hspace{6mm} [CONDITION heavy rain showers] [CLOUD\_COVERAGE partly cloudy] & \\
         & & \hspace{3mm} \orange{]} & \\
         & & \red{]} & \\
         \hline
         \multicolumn{4}{c}{\textbf{Annotated Response}} \\
         \multicolumn{4}{p{15cm}}{\red{[CONTRAST} \orange{[INFORM\_1} \blue{[LOCATION} [CITY Parker ] \blue{]} is not expecting any [CONDITION\_NOT snow] \orange{]}, but \orange{[INFORM\_2}  \blue{[DATE\_TIME} [COLLOQUIAL today] \blue{]} there's a [PRECIP\_CHANCE\_SUMMARY very likely chance] of [CONDITION heavy rain showers] and it'll be [CLOUD\_COVERAGE partly cloudy ] \orange{]} \red{]} }\\
    \end{tabularx}}
    \caption{Example response, MR, and other metadata from our dataset}
    \label{tab:weatherexamplerow}
\end{table*}

%\begin{figure}
%  \caption{Tree-structured annotation for the response in Table \ref{tab:weatherexamplerow}}
%  \label{figure:mr}
%  \centering
%  \includegraphics[width=0.48\textwidth]{images/weatherdataexample.png}
%\end{figure}
% \comment{Jinfeng: don't think this figure is very informative, as 1) the tree structures have been extensively stated before and are easy to recover from indented/colored MRs; 2) the figure itself is too small/blur to view.}

\subsection{Enriched E2E Dataset}\label{sec:e2edata}
We also used heuristic techniques to convert the E2E dataset to use tree-structured MRs. We used the output of Juraska et al.'s \citeyearpar{Slug2Slug} tagger to find a character within each slot in the flat MR, and automatically adjusted these to correspond to a token boundary if they didn't already. We then used the Viterbi segmentations from the model released by \citet{HarvardLearningTemplates} to get spans corresponding to each argument. Finally, we used the Berkeley neural parser \citep{BerkeleyNeuralParser} to identify spans coordinated by \textit{but}, and added \textsc{Contrast} relations as parents of the coordinated arguments. We added \textsc{Join} based on sentence boundaries.  An interesting direction for future research would be to extend Wiseman et al.'s methods to induce tree structures directly. In the final dataset we obtained (\texttildelow 51K examples), \texttildelow24K examples (47\%) contain \textsc{Join}, while 2237 (4.3\%) contain \textsc{Contrast}.

% this is already mentioned in the intro
%\footnote{We will release our version of the E2E dataset with this paper if accepted}

\section{Model}\label{sec:model}

\subsection{Seq2Seq with Linearized Trees}\label{subsec:delex}

In this work, we use a standard Seq2Seq model with attention \citep{seq2seqorig,bahdanauattention}, implemented in the \texttt{fairseq-py} repository \citep{fairseq}. The encoder and decoder are both Long Short-Term Memory (LSTM) -based \citep{lstmorig} and the decoder uses beam search for generation. The input to the model is a linearized representation of the tree-structured MR, and the output is a linearized tree-structured representation of the annotated response (see Table~\ref{tab:weatherexamplerow}).
% (where the tree structure is derived at training time from the annotated spans in the dataset). 
This means that in addition to predicting tokens for the surface realization of the response, the model must also predict non-terminals (dialog/discourse relations and arguments) to indicate the start or end of each span. One advantage of predicting a tree structure is that the model has supervision on the alignment between the MR and the response. Additionally, this predicted tree structure can be used to help verify the correctness of the predicted response; we leverage this for our constrained decoding approach described next. We also \textbf{delexicalized} tokens in the response that correspond to sparse entities, like names in the E2E dataset and temperatures in the weather dataset (see Appendix \ref{sec:datapreproc}).
%in the next section. 

\begin{figure}
%   \centering
  \includegraphics[width=0.43\textwidth]{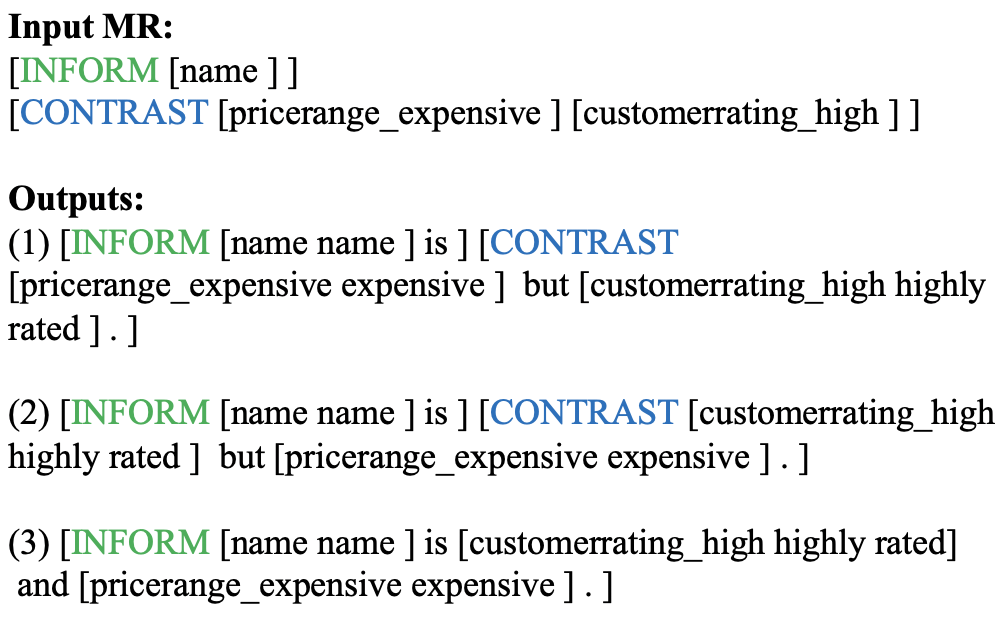}   \caption{Examples of constraint checking.   (1) and (2) are valid outputs. (3) fails to meet tree constraints since the \textsc{Contrast} node is not present and the \textsc{Inform} node has illegal children \texttt{customerrating} and \texttt{pricerange}.}   \label{figure:constraint_checking}
  \vspace{-0.5cm}
%In (2), the order of children of CONTRAST node is reversed. 
\end{figure}

\subsection{Constrained Decoding}\label{subsec:constrained}

As described above, the output structure predicted by the model forms a tree that should correspond neatly to the input MR, barring some instances of ellipses (as with the \texttt{date\_time} argument in Table \ref{tab:weatherexamplerow}).\footnote{A top-level \textsc{Join} is automatically added when necessary to create a single-rooted structure.}  Thus, the input MR can be seen as a constraint on the semantic correctness of the prediction; if the predicted structure doesn't match the MR, the prediction is incorrect and can be rejected. Figure~\ref{figure:constraint_checking} illustrates such ideas.

%We leverage this idea during beam search to help ensure that only correct candidates are generated by the model. 
%Specifically, our algorithm checks every node in the MR recursively, ensuring that all and only its children are present. 

Our beam search algorithm works as follows.\footnote{Pseudocode is given in the supplementary material.} First, the input tree is scanned to identify groups of two or more nodes that have the same value, so that ellipsis can be enabled by optionally allowing just one node in each group. Then, as the tree structure is incrementally decoded, non-terminals are checked against the input tree for validity.  When an opening bracket token (e.g., \texttt{[name}) is generated, it is not accepted if it isn't a child of the current parent node in the input tree, or has already been generated in the current subtree, thereby preventing repetition and hallucination of arguments or acts. When a closing bracket token \texttt{]} or an end-of-sentence (EOS) token is generated, it is  accepted only if all children of the current parent are covered either directly or through ellipsis, thus ensuring that all children of every node are generated. After each timestep of the beam search, the scores of candidates that violate tree constraints are masked so that they do not proceed forward. By removing candidates that violate the constraints early in the beam search, we allow the decoder to explore more hypotheses.

Checking these constraints and tracking coverage requires an alignment between the output and input MRs.  While the children of \textsc{Join} nodes are required to appear in order, child nodes of other discourse relations and dialogue acts can appear in any order, and thus the corresponding input non-terminal is not always uniquely identifiable when an output non-terminal is opened. For this reason, a set of possible alignments is maintained.  In particular, when accepting a non-terminal, all possible nodes in the input that it may correspond to are identified and a state is maintained for each possibility. Open states whose constraints are violated are removed from tracking, and a non-terminal is not accepted when no more open states are left. Though in principle the number of open states could grow large, empirically any alignment non-determinism is quickly resolved.

Note that although the algorithm ensures that the output tree structure is compatible with the input structure, it turns out that the model can still occasionally hallucinate content: since the neural model allows all possible token sequences in principle, it sometimes generates word sequences that express a hallucinated slot by simply skipping over the disallowed slot annotation---thereby bypassing the constraints---especially when given an unusual input. These cases are discussed further below.

\section{Experiments}\label{sec:expt}
In this section, we first describe our baselines, metrics, and implementation details, followed by experimental results and analyses.

\subsection{Experimental Setup}

%\par \textbf{Baselines.}
\paragraph{Baselines}
 We consider a few Seq2Seq-based baselines in our experiments (we use the open fairseq implementation~\citep{fairseq} for all our experiments). All models use an LSTM-based encoder and decoder, with attention.
\begin{description}[leftmargin=*,noitemsep]
\item[\textsc{S2S-Flat}] The input is a flat MR (for the E2E dataset, this is equivalent to the original form of the data; for weather, we remove all discourse relations and treat all dialog acts as a single large MR). The output is the raw delexicalized response.
\item[\textsc{S2S-Token}] Following \citet{Reed-et-al:W18-6535}, we add three tokens in the beginning of flat input MR (same as \textsc{S2S-Flat}) to indicate the number of contrasts, joins and number of sentences (dialog acts) to be generated.\footnote{Reed et al. only report results on controlling \textsc{Contrast} using an augmented training set, precluding direct comparison to their results.} The output is the raw delexicalized response. 
%\item[\textsc{Seq2Seq-Token}]: In the model proposed by \citet{reed2018}, the encoder sees a token indicating either contrast or the number of sentences. Since the authors were primarily interested in these effects independently, they did not experiment with a model that combines these forms of supervision. In order to ensure a fair contrast with our models, which can perform both contrast and sentence structure planning, we use a Seq2Seq model that sees tokens indicating contrast, the number of sentences (for the E2E dataset) or the number of dialog acts and discourse relations (for our weather dataset). Aside from these tokens, the input is a flat MR, and the output is the raw delexicalized response. 

\item[\textsc{S2S-Tree}] Same architecture as \textsc{S2S-Flat}, but the input and output for this model are the  linearized tree-structured MR and the tree-structured response respectively.
% \textbf{\textsc{Seq2Seq-Tree-Reranked}}: Same architecture and input/output representations as \textsc{Seq2Seq-Tree}. However, the final output is chosen by reranking all the candidates on the model's beam by tree accuracy (i.e. we pick the first candidate the matches the MR tree).\\
\item[\textsc{S2S-Constr}] Our proposed model. It has the same architecture as \textsc{S2S-Tree}, but decoding during beam search is constrained, as described in Section \ref{subsec:constrained}.
\end{description}

%\par \textbf{Data preprocessing.} 
\paragraph{Data preprocessing} 
%In data preparation, we use the reference-MR alignments to \textbf{filter} MRs and remove any information (arguments or dialog/discourse relations) that are not expressed in the reference or elided for redundancy. 
%We still keep arguments (with same values) that can be aggregated across dialog acts. This ensures our primary goal of controllability but still containing necessary discourse-level structuring, since the model no longer has to learn content selection.
%
In the input MR, all arguments within each dialog act are ordered alphabetically, to ensure a consistent ordering across examples. We also use alignments between the reference and the MR to \textbf{filter} information (arguments or dialog acts/discourse relations) that are not expressed in the reference; however, we ensure that any arguments that occur multiple times in the MR, but are elided in the reference for redundancy, are still preserved in the MR. This ensures that the model doesn't have to learn content selection, while still achieving our primary goal of discourse structure control. 

The inputs to \textsc{S2S-FLAT} and \textsc{S2S-Token} are prepared by removing all dialog act and discourse information in the linearized MR, and numbering arguments corresponding to the dialog act they belong in. Global order of dialog acts is preserved such that arguments of the first act occur before those arguments in the following acts, but arguments within a dialog act are ordered alphabetically. 
% An additional challenge of the weather dataset is referring expression generation (REG) for date/time arguments; for the purposes of our experiments, we assume that the model is provided with the correct surface form to use, and leave explorations of REG to future work.
% \comment{Jinfeng: don't think we need to talk about REG in this paper}

%\par \textbf{Metrics.} 
\paragraph{Metrics} 
We consider \emph{automatic} and \emph{human evaluation} metrics for our model. Automatic metrics are evaluated on the raw model predictions (which have delexicalized fields, like \texttt{temp\_low}):

\begin{itemize}[noitemsep, nolistsep, leftmargin=*]
    \item \textbf{Tree accuracy} is a novel metric that we introduce for this problem. It measures whether the tree structure in the prediction matches that of the input MR exactly. We implemented our tree accuracy metric to account for grouping and ellipsis, and will release this implementation along with our dataset.
    \item \textbf{BLEU-4}~\citep{Bleu:2002} is a word-overlap metric commonly used for evaluating NLG systems. 
\end{itemize}

Due to the limitations of automatic metrics for NLG \citep{automaticisbad1,automaticisbad2}, we also performed human evaluation studies by asking annotators to evaluate the quality of responses produced by different models. Annotators provided \textbf{binary} ratings on the following dimensions:

\begin{itemize}[noitemsep,nolistsep, leftmargin=*]
    \item \textbf{Grammaticality}: Measures \textit{fluency} of the responses. Our evaluation guidelines included considerations for proper subject-verb agreement, word order, repetition, and grammatical completeness.
    \item \textbf{Correctness}: Measures \textit{semantic correctness} of the responses. Our guidelines included considerations for sentence structure, contrast, hallucinations (incorrectly included attributes), and missing attributes. We asked annotators to evaluate model predictions against the reference (rather than the MR --- see Appendix \ref{sec:eval}). 
    % Since we filter MRs to reflect the reference, the model prediction should not vary in content from the reference. Even so, there is a risk that the model is unfairly penalized when evaluated against a reference that adds some other information not captured by the MR. We preferred this approach despite this risk, since we found that it greatly improves annotation speed and agreement rates \todo{maybe we should to cite our \#s on this, or provide more details in appendix}.
    % \comment{Jinfeng: again, not sure these are that important when space is scarce. Feel free to add it back if you still like to have it}
\end{itemize}

\subsection{Constrained Decoding Analysis}\label{subsec:constrdecexpts}

We trained each of the models described above on the weather dataset and the E2E dataset, and evaluated automatic metrics on the test set.\footnote{We used the scripts provided at \url{https://github.com/tuetschek/e2e-metrics} by the E2E organizers for evaluating both the E2E and the weather models.} In the E2E test set, each flat MR has multiple references (and therefore multiple compositional MRs). When computing BLEU scores for the token, tree, and constrained models, we generated one hypothesis for each of the compositional MRs for a single flat MR, and chose the hypothesis with the highest score against all references for that flat MR. We then computed corpus BLEU using these hypotheses. While this isn't an entirely fair way to evaluate these models against the E2E systems, it serves as a sanity check to validate that generation models provided with more semantic information about the references can achieve better BLEU scores against them. For both E2E and weather, we also filtered out, from all model computations, any examples where \textsc{S2S-Constr} failed to generate a valid response (\ref{subsec:results}).

For human evaluation, we show an overall correctness measure \textbf{\texttt{Corr}} measured on the full test sets, as well as \textbf{\texttt{Disc}}, measured on a more challenging subset of the test set that we selected. For the E2E dataset, we chose examples that contained contrasts by identifying references with a \textit{but} (230 total). For the weather dataset, we chose 400 examples where the MR has at least one \textsc{Contrast} or \textsc{Justify}. We also included test examples with argument type combinations previously unseen in the training set (313 total); we expect these to be challenging for all models, and in particular for the flat model, which has to infer the right discourse relation for new combinations of arguments. 

\begin{table*}[!]
  \small
  \begin{center}
    %\begin{tabularx}{1.042\linewidth}{|l|ccccc|ccccc|}
    \begin{tabularx}{1.0\linewidth}{l|ccccc|ccccc}
    \hline 
     \textbf{Model} & %\multicolumn{5}{c|}{\textbf{E2E}} & \multicolumn{5}{c|}{\textbf{Weather}}  \\ 
     \multicolumn{5}{c|}{\textbf{E2E}} & \multicolumn{5}{c}{\textbf{Weather}}  \\ 
     \textbf{Metric} & \textbf{BLEU} & \textbf{TreeAcc} & \textbf{Gram} & \textbf{Corr} & \textbf{Disc} & \textbf{BLEU} & \textbf{TreeAcc} & \textbf{Gram}  & \textbf{Corr} & \textbf{Disc} \\
     \hline
     \textbf{\textsc{S2S-Flat}}  & 0.6360 & - & 94.03 & 63.85 & 30.87 & 0.7455 & - & 98.77 & 77.09 & 79.04 \\
     \textbf{\textsc{S2S-Token}} & 0.7441\textdaggerdbl & -  & 92.29 & 69.02\textdagger & 42.29\textdagger & 0.7493\textsuperscript{*} & - & 96.7 & 81.56\textdagger & 83.93\textdagger \\
     \textbf{\textsc{S2S-Tree}}  & 0.7458\textdaggerdbl & 94.86  & 93.59 & 83.85\textdagger & 54.35\textdagger & 0.7612\textsuperscript{*} & 92.5 & 95.26 & 87.61\textdagger  & 85.97\textdagger \\
     \hline 
     \textbf{\textsc{S2S-Constr}} & \textbf{0.7469}\textdaggerdbl & \textbf{99.25} & \textbf{94.33} & \textbf{85.89}\textdagger & \textbf{66.09}\textdagger & \textbf{0.7660}\textsuperscript{*} &  \textbf{96.92} & 95.30 & \textbf{91.82}\textdagger & \textbf{93.44}\textdagger \\
    \hline
    \end{tabularx}
  \end{center}
   \caption{Automatic and human evaluated metrics on E2E and Weather datasets. All metrics other than  BLEU are percentages. \texttt{Corr} and \texttt{Disc} are the \% of examples for which the model prediction was judged by humans as semantically correct; \texttt{Disc} is measured on a challenging subset of \texttt{Corr}. \textsuperscript{*} indicates BLEU scores that are statistically significant ($p<0.01$) compared to \textit{all} baselines for that model. \textdaggerdbl indicates statistically significant BLEU scores ($p < 0.01$ ) compared to \textsc{S2S-Flat}. \textdagger indicates human-evaluated correctness scores that are statistically significant ($p<0.05$), using McNemar's chi-squared test, compared to \textit{all} baselines for that model.}
     \label{tab:constrainedresults}%
     \vspace{-0.3cm}
\end{table*}%

\subsection{Results}\label{subsec:results}

Table \ref{tab:constrainedresults} shows the results of this experiment. On both the E2E and weather datasets, \textsc{S2S-Constr} improves tree accuracy significantly (using McNemar's chi-squared test) over \textsc{S2S-Tree}. Human evaluation metrics also show that models that are aware of the tree-structured MR (\textsc{S2S-Tree} and \textsc{S2S-Constr}) perform significantly better on correctness measures than \textsc{S2S-Token}, which is only aware of the presence or absence of discourse relations, and significantly better than \textsc{S2S-Flat}, which has no awareness of the structure. The gap is larger on \textbf{\texttt{Disc}}: the flat model gets only 31\% of the challenging cases correct on the E2E dataset, while the constrained model's accuracy is more than twice that. A similar gap is evident in the weather dataset. 
Further, \textsc{S2S-Constr}, \textsc{S2S-Tree}, and \textsc{S2S-Token} all show significant improvements in BLEU over the flat baseline. These systems also outperform the E2E baseline \textsc{TGen} \citep{tgenbaseline} and the challenge winner \textsc{SLUG} \citep{Slug2Slug} on BLEU (0.6519 and 0.6693 respectively, from \citet{e2echallengeoverview}) and diversity metrics (Section~\ref{subsec:diversity}).  We note that for the E2E dataset, the BLEU score increases observed with the tree-based models are not statistically significant compared to \textsc{S2S-Token}.  We think this may be partly because many discourse patterns are correlated with the flat MR structure in the E2E dataset (e.g. \texttt{family-friendly} and \texttt{highly rated} are frequently \textsc{Contrast}ed). By contrast, BLEU score increases are statistically significant for all models on our weather dataset. Also, \textsc{S2S-Constr} fails to generate any valid candidates for \texttildelow1.5\% of the weather test examples. In most of these cases, the model \textit{stutters}, i.e. produces degenerate output like ``will be be be \textellipsis''. We suspect that in these cases, the imposed decoding constraints cause the Seq2Seq decoder to get stuck in a pseudoterminal state.

Grammaticality seems to drop slightly for the tree-based models on the weather dataset, but not on the E2E dataset. One hypothesis from this and the correctness numbers is that the flat models generate more generic (and therefore grammatical), but also incorrect, responses, compared to the tree-based models.  We also note that there's a noticeable gap in the E2E dataset between tree accuracy and the correctness numbers from human evaluation. We analyzed 35 examples where our tree accuracy metric disagreed with human evaluation, and found 22 (63\%) cases where the compositional MR was missing information in the reference, seemingly due to noise in our automatic annotation process (Section \ref{sec:e2edata}). We also identified 6 cases (17\%) of annotator confusion (for example whether \textit{between \pounds20-30} implies the same meaning as \textit{moderately priced}), sometimes caused by noisy references that contained additional information. The remaining examples all contained legitimate model errors, like content hallucination, or a wrong slot being produced despite a correct non-terminal. One future direction to get more reliable metrics would be to improve the automatic annotation process in Section \ref{sec:e2edata} to eliminate noise and flag noisy references. Further experimentation is described in Appendix \ref{sec:additionalexp}.

% Finally, constrained decoding appears to have had at most a small negative impact on grammaticality through human evaluation. 
% \comment{Jinfeng: constrained decoding actually improves over Seq2seq-tree on grammaticality.}

\begin{figure}
 \centering
 \includegraphics[width=0.48\textwidth]{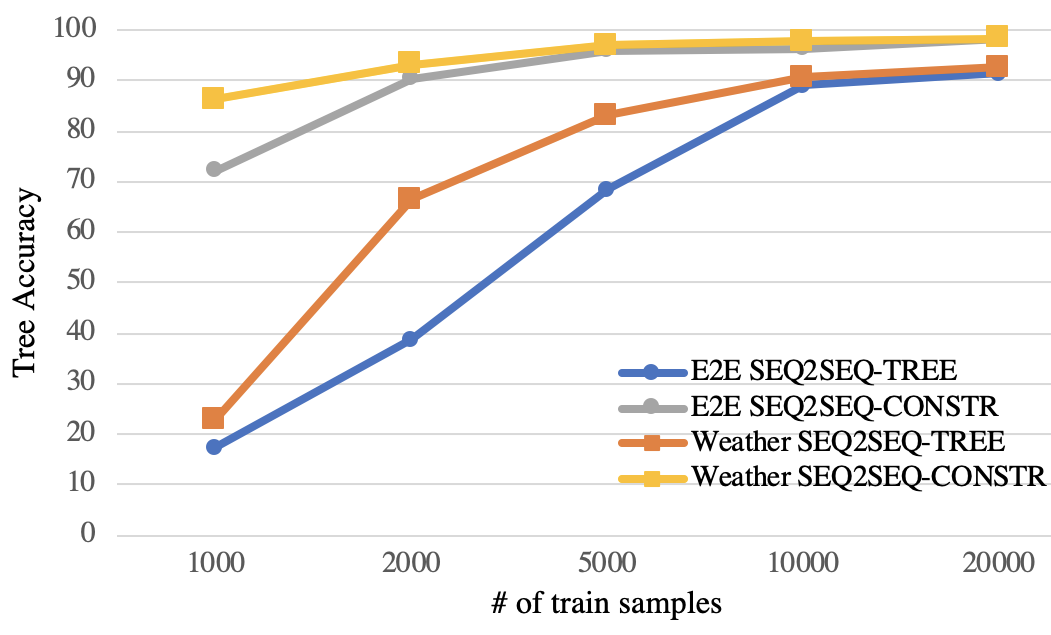}
 \caption{Performance of models on test set for varying number of samples in train set.}
 \label{figure:train_set_number_samples}
 \vspace{-0.6cm}
\end{figure}

\subsection{Diversity Metrics}\label{subsec:diversity}
We report the diversity metrics used for evaluating E2E challenge submissions in \citet{e2echallengeoverview} (\# unique tokens, \# unique trigrams, Shannon token entropy \citep[p.61ff.]{Manning:1999:FSN:311445}, conditional bigram entropy \citep[p.63ff.]{Manning:1999:FSN:311445}). Table \ref{tab:diversityresults} shows these numbers, as compared against a few of the E2E participating systems,  \textsc{TGen}, \textsc{SLUG}, and \textsc{ADAPT} \citep{adaptmodel}. All of the models with enriched semantic representations --- \textsc{S2S-Token}, \textsc{S2S-Tree}, and \textsc{S2S-Constr} --- show higher diversity than neural baselines without diversity considerations. Combined with our improved BLEU scores, this seems to indicate that adding discourse relation information to input MRs can increase diversity, without incurring losses on automatic metrics (as is the case with the diversity-promoting \textsc{ADAPT} system).
\begin{table*}[!]
  \small
  \begin{center}
    \begin{tabular}{|l|p{1.25cm}p{1.25cm}p{1.25cm}p{1.25cm}|}
    \hline 
     \textbf{Model} & \textbf{Unique tokens} & \textbf{Unique trigrams} & \textbf{Shannon entropy} & \textbf{Cond. entropy bigrams}\\
     \hline
     \rowcolor{Gray}
     \textbf{\textsc{TGen}} & 83 & 597 & 5.41 & 1.32 \\ 
     \rowcolor{Gray}
     \textbf{\textsc{SLUG}} & 74 & 507 & 5.35 & 1.13 \\ 
     \rowcolor{Gray}
     \textbf{\textsc{ADAPT}} & 455 & 3567 & 6.18 & 2.09 \\ 
     \hline
     \textbf{\textsc{S2S-Token}} & 137 & 1147  & 5.86 & 1.71\\
     \textbf{\textsc{S2S-Tree}}  & 134 & 1030  & 5.85 & 1.65 \\
     \hline 
     \textbf{\textsc{S2S-Constr}} & 134 & 1128  & 5.86 & 1.71 \\
%     TGen \citep{tgenbaseline} & 0.6593 &  &  &  &  &  &  & N/A & &  \\
%     \citet{reed2018} (our implementation) & 0.6376 & & & & & & & N/A & &  \\
%     Seq2seq~\citep{fairseq} & 0.6395 & & 90.16\% & & & 0.92 & & 94.25\%  & & \\
%     Seq2seq + tree accuracy-based reranking & & & & & & & & 95.44 & & \% \\
%     \hline
%     Seq2seq + constrained decoding & 0.4713 & & \textbf{96.98}\% & & & 0.91 & & \textbf{96.92\%} & \\
    \hline
    \end{tabular}
  \end{center}
   \caption{E2E dataset diversity metrics. Rows in gray correspond to metrics that we cite from \citet{e2echallengeoverview}.}
     \label{tab:diversityresults}%
\end{table*}

\subsection{Data Efficiency and Generalizability}

We measured tree accuracy on the full E2E and weather test sets by varying the number of training samples for \textsc{S2S-Tree} and \textsc{S2S-Constr} (Figure \ref{figure:train_set_number_samples}). \textsc{S2S-Constr} achieves more than 90\% tree accuracy with just 2K samples and more than 95\% with 5K samples on both datasets, suggesting that constrained decoding can help achieve superior performance with much less data.

\begin{figure}
 \centering
 \includegraphics[width=0.35\textwidth]{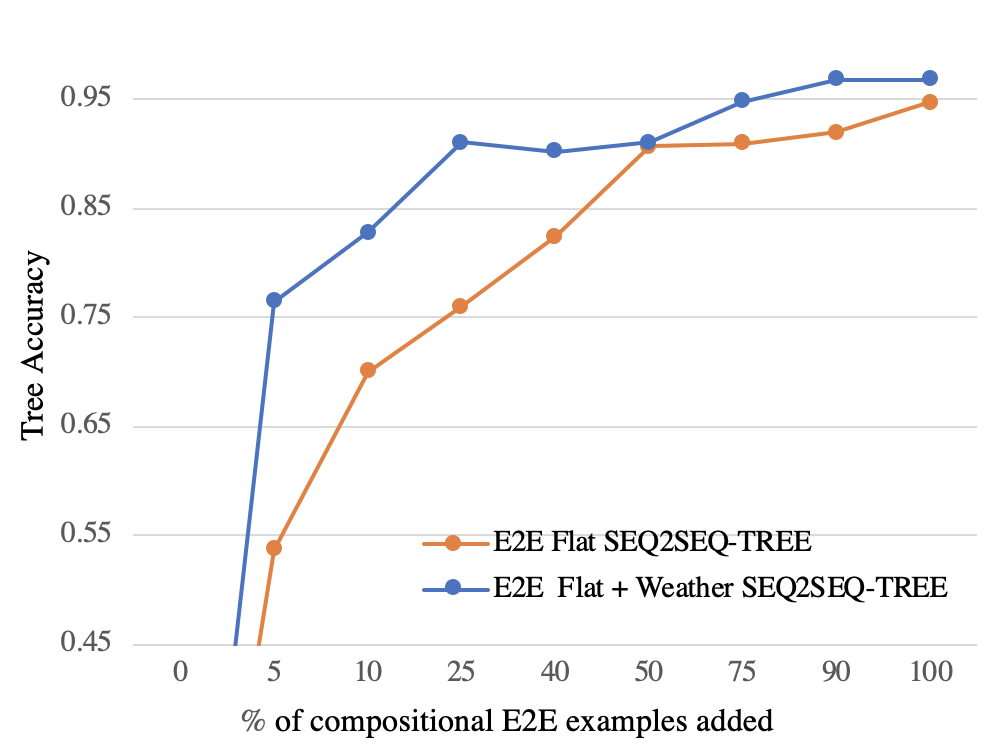}
 \caption{Performance of \textsc{S2S-Tree} models trained on E2E flat data, and flat E2E + full weather dataset, with a fraction of composition E2E.}
 \label{figure:domain_adaptation}
 \vspace{-0.3cm}
\end{figure}

Meanwhile, we also investigated the extent to which tree-structured MRs could allow models to generalize to compositional semantics (Figure \ref{figure:domain_adaptation}). We first split the complete E2E training set into flat and compositional examples (26896 vs.\ 24530), where flat examples don't contain any discourse relations. Next, we trained a model on the full weather dataset and flat E2E data, gradually added more compositional E2E samples to the training set, and checked the model's accuracy on a test set with only compositional examples. Without any compositional E2E examples, both models fail to produce any valid sequences (not pictured). However, when just 5\% of the compositional examples are added to the training data, the \textsc{E2E-Weather} model gets a tree accuracy of 76\%, while the model trained on E2E only gets 53.72\%. The final \textsc{E2E-Weather} model also has higher overall accuracy than the E2E-only model. This shows that learned discourse relations can be leveraged for domain adaptation.

%The figure also shows that constrained decoding makes the model much more data efficient.

\section{Related Work}\label{sec:related} 

Reed et al.'s \citeyearpar{Reed-et-al:W18-6535} approach to enriching the input, discussed earlier, is the most closely related work to ours. A more recent work by \citet{naacl2019separateplanning} also focuses on exercising more control over input structures through sentence plans; however, their work doesn't touch on discourse relations or constrained decoding. \citet{puduppully2018data} builds a modular end-to-end neural architecture that performs content planning in addition to realization, although they focus on generating text from structured tables, and don't consider discourse structure. 

Also related is Kiddon et al.'s \citeyearpar{D16-1032} neural checklist model, which tracks the coverage of an input list of ingredients when generating recipes. Our constrained decoding approach goes beyond covering a simple list by enforcing constraints on ordering and grouping of tree structures, but theirs takes coverage into account during model training.  A more direct inspiration for our approach is the way coverage has been traditionally tracked in grammar-based surface realization \citep{Shieber:1988,kay:1996:ACL,Carroll-and-co:1999,Carroll-Oepen:2005,Nakanishi-Miyao-Tsujii:iwpt05,White2006,white-rajkumar:2009:EMNLP}.  Compared to our approach, grammar-based realizers can prevent hallucination entirely, though at the expense of developing an explicit grammar. Constrained decoding in MT \citep[i.a.]{Post2018FastLC} has been used to enforce the use of specific words in the output, rather than constraints on tree structures.  Also related are neural generators that take Abstract Meaning Representations (AMRs) as input \citep[i.a.]{konstas-EtAl:2017:Long} rather than flat inputs; these approaches, however, do not generate trees or use constrained decoding.

\section{Conclusions}

We show that using rich tree-structured meaning representations can improve expressiveness and semantic correctness in generation. We also propose a constrained decoding technique that leverages tree-structured MRs to exert precise control over the discourse structure and semantic correctness of the generated text. We release a challenging new dataset for the weather domain and an enriched E2E dataset that include tree-structured MRs. Our experiments show that constrained decoding, together with tree-structured MRs, can greatly improve semantic correctness as well as enhance data efficiency and generalizability. 

% domain adaptation actually not that strong?
%We also show that models trained on these representations perform very well at domain adaptation tasks.

% Human evaluation seems to indicate, however, that constrained decoding can still hallucinate content. This suggests the potential for using additional checks for semantic correctness. One such approach could involve using parsers to predict the tree structure of a given response, and compare that against the MR.
% \comment{Jinfeng: let's leave future work away for now}

% probably not space for this?
%The dataset that we release contains a good amount of complexity that could be explored in future work alongside the tree-structured MRs. It would be interesting to explore the correctness impact of conditioning on the user query.

\section*{Acknowledgments}

We thank Juraj Juraska for supplying the output of their E2E slot tagger.  We also thank Michael Auli, Alex Baevski and Mike Lewis for discussion.

%The acknowledgments should go immediately before the references.  Do
%not number the acknowledgments section. Do not include this section
%when submitting your paper for review. \\

\bibliography{acl2019}
\bibliographystyle{acl_natbib}

\clearpage
\appendix

\section{Weather Forecast Generation}
\label{sec:scengen}
For every example, we extracted the date range requested by the user, and generated artificial weather forecasts for that date range. We generated forecasts of different granularities (hourly or daily) depending on the date requested by the user. If the date that requested was less than 24 hours after the ``reference'' date in the synthetic user context, we generated hourly forecasts; otherwise, we generated the required number of daily forecasts. To generate forecasts, we selected reasonable mean, standard deviation, min, and max values for temperature and cloud coverage, and used these to sample temperatures for every point in the date range. We also selected random sunrise and sunset times for each day present in the range. We picked values that seemed reasonable, but didn't try too hard to get precise values, since our focus was more on using the forecasts to create complex MRs. After sampling temperatures and cloud coverage amounts for each range, we randomly chose other attributes to include, conditioned on the values of the temperatures and cloud coverage, like precipitation chance, wind speed summary, and other rarer conditions like fog.

\section{Tree-Structured Weather MR Creation}
\label{sec:mrcreation}
\begin{enumerate}[leftmargin=*]
    \item Errors: We added \texttt{ERROR} dialog acts whenever the user query contained a weather request for a date too far in the future. We also chose locations to treat as ``unknown'' randomly, thus adding errors for locations unknown to the system. These ERROR acts are interesting because they capture domain-specific information about the nature and cause of errors, and can potentially be learned across domains. Additionally, including ERROR acts creates scope for interesting responses like ``I'm sorry, I don't know where that is. But right now in [user's default location], it's sunny ...''.
    \item Aggregation: We identified dates that had similar weather attributes (precipitation, cloud coverage, etc.) and created \texttt{INFORM} dialog acts that expressed information regarding each date. We then grouped these acts together using a \texttt{JOIN} discourse relation.
    \item Contrast: We identified attributes that were in opposition (``cloudy'' vs. ``sunny'') and added a parent \textsc{Contrast} discourse relation to any such dialog acts. We also contrasted related attributes wherever possible; e.g. the cloud coverage value ``sunny'' can be contrasted with both ``cloudy'' and the precipitation type ``rain''.
    \item Yes/no questions: Whenever the user query was a boolean one (``Will it rain tomorrow''), we added \texttt{YES} or \texttt{NO} dialog acts as appropriate.
    \item Justifications/Recommendations: Whenever the user query mentioned an attire or activity (``Should I wear a raincoat?''), we assumed that the MR should communicate a recommendation as well as a justification for it (``No, you don't need to wear one, it looks like it'll be sunny all day''). In these cases, we added a \texttt{RECOMMEND} dialog act, and an \texttt{INFORM} dialog act that provides the justification for the recommendation. We added a parent \texttt{JUSTIFY} discourse relation to these acts, treating the recommendation as the nucleus and the \texttt{INFORM} as the satellite of the justification.
\end{enumerate}

\section{Dataset Creation Quality}
As mentioned in \ref{sec:data}, we asked annotators to evaluate collected responses, and used these to filter out noisy references and annotations from our final dataset. The ratings were made on a 1-5 scale and double annotated, and we filtered out 3,404 examples (out of a total 37,162) that had a score less than 3 on any of the four dimensions: fluency, correctness, naturalness, annotation correctness. 

\section{Data Preprocessing}
\label{sec:datapreproc}
Infinitely-valued arguments such as names of restaurants, dates, times, and locations such as cities, states are delexicalized (value is replaced by placeholder tokens) in both the input and output of models. This was done following the approach taken by several of the systems in the E2E challenge \citep{tgenbaseline,Slug2Slug,e2echallengeoverview}. The reasoning behind this is that the values of such arguments are often inserted verbatim in the response text, and therefore do not affect the final surface form realization. Replacing these arguments in both the input and output reduces the vocabulary size and prevents sparsity issues. (A copy mechanism, such as the one introduced in \citet{copynetworks}, can be used to address this, though we did not explore this approach in this work.) The full list of arguments for which we performed delexicalization is:
\begin{enumerate}
    \item Numerical arguments: temperature-related arguments, precipitation chance, day, month, year (for dates). 
    \item Named entities: restaurant name (E2E), landmark (E2E), city, region, country, weekday (for dates)
\end{enumerate}

\section{Additional Experiments}
\label{sec:additionalexp}
We also experimented with a reranked \textsc{S2S-Tree} in which the beam search candidates are reranked for tree accuracy. This yields a tree accuracy of 97.6\% and 95.4\% on E2E and weather.

We trained a Recurrent Neural Network Grammar (RNNG) to tag slots in the prediction of \textsc{S2S-Constr} in order to filter out hallucinations. The correctness on filtered test sets rose from 85.89\% to 87.44\% for E2E, and from 91.82\% to 93.84\% on weather.

\section{Human Evaluation of Models}\label{sec:eval}
When asking annotators to rate the models on correctness, we asked them to rate the response by comparing it against the reference, rather than against the MR. This adds the risk that annotators are confused by noisy references, but we found that it increased annotation speed and agreement rates significantly over evaluating against the MR directly. This is also because our MRs are tree-structured and can be hard to read. We performed double-annotation with a resolution round. 
\textbf{Automatic rejection:} When analyzing evaluation results, we found that it was fairly easy to miss the absence of a contrast or a justification in our weather dataset, especially since our dataset is so large. As a result, annotators were marking several incorrect cases as correct. To address this issue, we automatically marked as incorrect any examples where the MR had a \textsc{Contrast} but the response lacked any contrastive tokens, or where the MR has a \textsc{Justify} but the response lacked any clear markers of a justification. This eliminated noise from  ~2.8\% of all responses.

\section{Model Training Details}
We used the same seq2seq model from the \textsc{S2S-Flat} baseline for our constrained decoding experiments, which used 300-dimensional GloVe word embeddings~\cite{pennington2014glove}, a dropout rate of 0.2 \citep{srivastava2014dropout}, and hidden dimension of 128 in both the encoder and the decoder. We used the Adam optimizer \citep{Kingma2015AdamAM} with a learning rate of 0.002 to train the seq2seq model. The learning rate is reduced by a factor of 5 if the validation loss stops decreasing. Beam size is set to 10.

\newpage
\section{Constrained Decoding Algorithm} \label{subsec:pseudcode} 
\label{sec:supplemental}
% Submissions may include non-readable supplementary material used in the work and described in the paper. Any accompanying software and/or data should include licenses and documentation of research review as appropriate. Supplementary material may report preprocessing decisions, model parameters, and other details necessary for the replication of the experiments reported in the paper. Seemingly small preprocessing decisions can sometimes make a large difference in performance, so it is crucial to record such decisions to precisely characterize state-of-the-art methods. 

% Nonetheless, supplementary material should be supplementary (rather
% than central) to the paper. \textbf{Submissions that misuse the supplementary 
% material may be rejected without review.}
% Supplementary material may include explanations or details
% of proofs or derivations that do not fit into the paper, lists of
% features or feature templates, sample inputs and outputs for a system,
% pseudo-code or source code, and data. (Source code and data should
% be separate uploads, rather than part of the paper).

% The paper should not rely on the supplementary material: while the paper
% may refer to and cite the supplementary material and the supplementary material will be available to the
% reviewers, they will not be asked to review the
% supplementary material.

\begin{lrbox}{\mybox}%
\begin{lstlisting}%[basicstyle=\small, language=Python, caption=Pseudo-code for constrained decoding as employed by \textsc{S2S-Constr}]

def build_constraints(MR):
    # nodes in MR are numbered from 0 to n in order of their
    # discovery in depth-first-search.
    # example, for MR: [JOIN [INFORM [A ] [B ] ] [INFORM [B ] [D ]]]
    # ids: JOIN: 0, INFORM: 1, A: 2, B: 3, INFORM: 4, B: 5, D: 6
    for node in MR:
        parent_map[node.id] = node.parent
        children_map[node.id] = node.children
        # map from non-terminal to all node ids of the non-terminal
        # eg: INFORM -> {1, 4} in case of example MR above
        valid_non_terminal_nodes[node.non_terminal].add(node.id)
    # map from node id to nodes that can cover it through ellipsis
    # example, for above MR: {3: {3, 5}, 5: {3, 5}}
    ellipsis_options = compute_ellipsis_options(MR)
    init_state.parent = -1  # current parent
    init_state.coverage = {}  # tracks node ids encountered till now
    # track nodes that have been covered through ellipsis
    init_states.elided_nodes = {}
    states = [init_state]  # list of open states

def children_covered(state, node):
    # returns true if all nodes have covered either
    # directly or through ellipsis
    missing_children = children_map[state.parent] - state.coverage
    for missing_child in missing_children:
        if (ellipsis_options[missing_child] 
               - state.elided_nodes) is empty:
            # nodes that have been elided themselves 
            # can't cover other nodes through ellipsis
            return False
    return True

def accept_token(states, next_token):
    # move states one time-step forward by accepting next_token
    # returns False if next_token cannot be accepted by any state
    if not next_token.startswith("[") or next_token != "]":
        # only non-terminal tokens need to be checked
        return True
    updated_states = []
    for state in states:
        if next_token.startswith("["):
            for candidate in valid_non_terminal_nodes[next_token]:
                if candidate in children_map[state.parent] 
                      and candidate not in state.coverage:
                    # create a new state for each valid candidate
                    new_state = copy(state)
                    new_state.parent = candidate
                    new_state.coverage.add(candidate)
                    updated_states.append(new_state)
        elif next_token == "]" 
                and children_covered(state, state.parent):
            # accept closing brace for current node and 
            # move states up a level in tree
            new_state = copy(state)
            new_state.parent = parent_map[state.parent]
            missing_children = 
                  children_map[state.parent] - state.coverage
            # if we're accepting a closing node with missing children,
            # then all of them must be getting elided
            new_state.elided_nodes.add(missing_children)
            updated_states.append(update(new_state, next_token))
    states = updated_states
    return len(states) > 0

def mask_score(score, states, next_token):
    if accept_token(states, next_token):
        return score
    else:
        return 0
\end{lstlisting}%
\end{lrbox}%
\begin{center}
\scalebox{0.5}{\usebox{\mybox}}
\end{center}

\end{document}